\documentclass[twocolumn,a4paper]{article}
\usepackage{nolta2019}
\usepackage{amsmath} %<--- Please uncomment this command if you use amsmath package.
\usepackage{txfonts}
\usepackage{algorithm}
\usepackage{graphicx}
\usepackage{algpseudocode}
\usepackage{cite}
\usepackage{multirow} 
\usepackage{ulem}

\begin{document}

\title{An Adaptive Stochastic Nesterov Accelerated Quasi Newton Method for Training RNNs}

\author{S. Indrapriyadarsini${}^\dag$ , Shahrzad Mahboubi${}^\S$, Hiroshi Ninomiya${}^\S$ and Hideki Asai${}^\ddag$}

\address{
\dag Graduate School of Integrated Science and Technology, Shizuoka University\\
\ddag Research Institute of Electronics, Shizuoka University\\
3-5-1 Johoku, Naka-ku, Hamamatsu, Shizuoka Prefecture 432-8011, Japan\\
\S Graduate School of Electrical and Information Engineering, Shonan Institute of Technology\\
1-1-25 Tsujido-nishikaigan, Fujisawa, Kanagawa Prefecture 251-8511, Japan\\
Email: s.indrapriyadarsini.17@shizuoka.ac.jp, 18T2012@sit.shonan-it.ac.jp, ninomiya@info.shonan-it.ac.jp, \\asai.hideki@shizuoka.ac.jp
}

\maketitle

\abstract
A common problem in training neural networks is the vanishing and/or exploding gradient problem which is more prominently seen in training of Recurrent Neural Networks (RNNs). Thus several algorithms have been proposed for training RNNs. This paper proposes a novel adaptive stochastic Nesterov accelerated quasi-Newton (aSNAQ) method for training RNNs. The proposed method aSNAQ is an accelerated method that uses the Nesterov's gradient term along with second order curvature information. The performance of the proposed method is evaluated in Tensorflow on benchmark sequence modeling problems.  The results show an improved performance while maintaining a low per-iteration cost and thus can be effectively used to train RNNs.
\endabstract
\vspace{-3mm}
\section{Introduction}\vspace{-2mm}
Recurrent Neural Networks (RNNs) are powerful sequence models \cite{goodfellow2016deep}. They are popularly used in solving pattern recognition and sequence modelling problems such as text generation, image captioning, machine translation, speech recognition, etc. However training of RNNs is very difficult mainly due to the vanishing and/or exploding gradient problem \cite{pascanu2013difficulty}. Hence several algorithms and architectures have been proposed to address the issues involved in training RNNs \cite{sutskever2013training}. Architectures such as Long Short-Term Memory (LSTM) and Gated Recurrent Units (GRU) have shown to be more resilient to the gradient issues compared to vanilla RNNs. Several other studies revolve around proposing algorithms that can be effectively used in training RNNs. Some of these methods propose the use of second order curvature information \cite{martens2011learning}. However, compared to first order methods, second order methods have a higher per-iteration cost.  Thus recent studies \cite{sutskever2013training,martens2011learning,le2015simple,keskar2016adaqn} propose algorithms that judiciously incorporates curvature information while taking the computation cost into consideration.  

In this paper, we propose a novel adaptive stochastic Nesterov Accelerated quasi-Newton (aSNAQ) method. The proposed method is similar to the framework of SQN \cite{byrd2016stochastic} and adaQN \cite{keskar2016adaqn} with some changes which are described in later sections.  This paper attempts to study the performance of the proposed algorithm for training RNNs in comparison with adaQN and popular first order methods Adam and Adagrad.
%They can also be applied to image classifications problems by decomposing the images into a series.  Recurrent Neural Networks (RNNs) are powerful sequence models. They are popularly used in solving pattern recognition and sequence model problems such a text generation, image captioning, machine translation, speech recognition, etc. However training of RNNs is very difficult mainly due to the vanishing and/or exploding gradient issue. Hence several algorithms and techniques have been proposed. Recent They can also be applied to image classifications problems by decomposing the images into a series.  
%\section{Formulation of Training and Gradient Based Training Methods}
\vspace{-3mm}
\section{Background}\vspace{-1.5mm}

%\subsection{Formulation of Training}
Training in neural networks is an iterative process in which the parameters are updated in order to minimize an objective function. Given a mini-batch ${X \subseteq T_r}$  with samples ${(x_p,d_p)_{p \in X}}$ drawn at random from the training set ${T_r}$ and error function ${E_p({\bf w} ;x_p,d_p)}$ parameterized by a vector ${{\bf w} \in \mathbb{R}^d}$,  the objective function is defined as 
\vspace{-1mm}
\begin{equation}\label{eq:obj} \underset{{\bf w} \in \mathbb{R}^d}{\text {min}} E({\bf w})= \frac {1}{b}{ \sum_{p \in X} E_p ({\bf w})}, 
\vspace{-1mm}
\end{equation}
 where ${b}={|X|}$, is the batch size. In gradient based methods, the objective function ${E({\bf w})}$ under consideration is minimized by the iterative formula 
\vspace{-1mm}
\begin{equation}\label{eq:1} 
{\bf w}_{k+1} = {\bf w}_k + {\bf v}_{k+1}.
\vspace{-1mm}
\end{equation} 
where ${\it k}$ is the iteration count and ${\bf v}_{k+1}$ is the  update vector, which is defined for each gradient algorithm. 

\vspace{-2mm}
\subsection{BFGS quasi-Newton method}
Quasi-Newton (QN) methods utilize the gradient of the objective function to result in superlinear quadratic convergence. The Broyden-Fletcher-Goldfarb-Shanon (BFGS) algorithm is one of the most popular quasi-Newton methods for unconstrained optimization \cite{nocedal2006}. The update vector of the QN method is given as
\vspace{-1mm}
\begin{equation}
{\bf v}_{k+1} =  \alpha_k {\bf {g}}_k,
\end{equation} 
\vspace{-5mm}
\begin{equation}
{\bf {g}}_k=-{\bf {H}}_k \nabla E({\bf w}_k).
\end{equation}
%\begin{equation}
%{\bf g}_k = -{\bf H}_k \nabla E({\bf w}_k).
%\end{equation}
The Hessian matrix ${\bf H}_k$ is symmetric positive definite and is iteratively approximated by the BFGS formula \cite{nocedal2006},
\begin{equation}\label{eq:5}
{\bf H}_{k+1}= ( {\bf I}- {\bf s}_k {\bf y}_k^{\rm T}/{{\bf y}_k^{\rm T} {\bf s}_k}){\bf H}_k({\bf I}- {\bf y}_k {\bf s}_k^{\rm T}/{{\bf y}_k^{\rm T} {\bf s}_k})+ {\bf s}_k {\bf s}_k^{\rm T}/{{\bf y}_k^{\rm T} {\bf s}_k},
\end{equation}
where ${\bf I}$ denotes identity matrix,
\begin{equation}
{\bf s}_k = {\bf w}_{k+1} - {\bf w}_k 
%~~{\rm and}~~ 
\end{equation}
\begin{equation}
{\bf y}_ k = \nabla E ( {\bf w}_{k+1} ) - \nabla E ({\bf w}_k).
\end{equation}
%As the scale of the neural network model increases, the cost of storing and updating the Hessian matrix ${\bf H}_k$ is expensive. 

\subsubsection{Limited Memory BFGS Method (LBFGS)}
\vspace{-2mm}
%One of the major issues with second order methods such as the BFGS quasi-Newton method is the computation and storage of the Hessian.
As the scale of the problem increases, the cost of computation and storage of the Hessian matrix becomes expensive. Limited memory scheme help reduce the cost considerably, especially in stochastic settings where the computations are based on small mini-batches of size ${b}$. In the limited memory LBFGS method, the Hessian matrix is defined by applying ${m_L}$ BFGS updates using only the last ${m_L}$ curvature pairs ${\{{\bf s}_k,{\bf y}_k\}}$, where $m_L$ denotes the memory size. The search direction ${\bf g}_k$ is evaluated using the two-loop recursion \cite{nocedal2006} as shown in Algorithm 1. %Thus, the computation and storage cost is significantly reduced. 

\iffalse
\subsubsection{Nesterov Accelerated Quasi-Newton }

The Nesterov's Accelerated Quasi-Newton (NAQ) ~\cite{ninomiya2017novel} method achieves faster convergence compared to the standard quasi-Newton methods by quadratic approximation of the objective function at ${\bf w}_k+\mu {\bf v}_k$ and by incorporating the Nesterov's accelerated gradient $\nabla E({\bf w}_k+\mu {\bf v}_k)$ in its Hessian update.  Therefore, the update vector of NAQ can be written as:
\begin{equation}
{\bf v}_{k+1} = \mu_k {\bf v}_k + \alpha_k {\bf {\hat g}}_k,
\end{equation} 
\begin{equation}
 {\bf {\hat g}}_k=-{\bf {\hat H}}_k \nabla E({\bf w}_k+\mu_k {\bf v}_k).
\end{equation}.
 The Hessian matrix ${\bf \hat H}_{k+1}$ is updated using

\begin{equation}\label{eq:12}
{\bf {\hat H}}_{k+1}= ( {\bf I}-{\bf p}_k {\bf q}_k^{\rm T}/{{\bf q}_k^{\rm T} {\bf p}_k}){\bf {\hat H}}_k({\bf I}- {\bf q}_k {\bf p}_k^{\rm T}/{{\bf q}_k^{\rm T} {\bf p}_k})+ {\bf p}_k {\bf p}_k^{\rm T}/{{\bf q}_k^{\rm T} {\bf p}_k},
\end{equation}
 \begin{equation}\label{eq:13}
{\bf p}_k = {\bf w}_{k+1} - ({\bf w}_k+ \mu{\bf v}_k) ~~{\rm and}~~{\bf q}_k = \nabla E ( {\bf w}_{k+1} ) - \nabla E ({\bf w}_k+ \mu {\bf v}_k).
\end{equation}
In the limited memory form, LNAQ~\cite{LNAQ_shah} uses the last {\it m} curvature pairs  for the Hessian calculation. The curvature pairs that are used incorporate the momemtum and Nesterov's accelerated gradient term, thus accelerating LBFGS. 
\fi
\vspace{-1mm}
\subsection{adaQN}
\vspace{-1mm}
adaQN is a recently proposed method which was shown to be suitable for training RNNs as well \cite{keskar2016adaqn}. It builds on the algorithmic framework of SQN \cite{byrd2016stochastic} by decoupling the iterate and update cycles. adaQN targets the vanishing/exploding gradient issue by initializing ${\bf H}_k^{(0)}$ in the two-loop recursion (Algorithm 1, step 7) based on the accumulated gradient information as shown below.
\vspace{-2.5mm}
\begin{equation}
[H_k^{(0)}]_{ii} = \frac{1}{\sqrt{ {\sum_{j=0}^{k} \nabla E({\bf w}_j)_i^2 } + \epsilon}} 
\vspace{-2mm}
\end{equation}
\vspace{-2mm}

 adaQN proposes the use of an accumulated Fisher Information matrix (aFIM) that stores at each iteration the ${\nabla E({\bf w}_k)}{\nabla E({\bf w}_k)^T }$ matrix in a FIFO memory buffer ${F}$ of size $m_F$. This is used in the computation of the ${\bf y}$ vector for Hessian approximation as   
\vspace{-3.5mm}
\begin{equation}
{\bf y}= \frac{1}{|F|}{ \sum_{i=1}^{|F|}{F}_i \cdot s}% {\rm ~~where~~{{F}_i = \nabla_i E({\bf w})}{\nabla_i E({\bf w})^T }}
\vspace{-3mm}
\end{equation}

where ${F}_i = {\nabla E({\bf w}_k)}{\nabla E({\bf w}_k)^T }$ and $|F|$ is the number of $F_i $ entries present in $F$.
In practice, ${\bf y}$  is computed without explicitly constructing the ${\nabla E({\bf w}_k)}{\nabla E({\bf w}_k)^T }$ matrix. Hence it is sufficient to just store ${\nabla E({\bf w}_k)}$. %The Fisher Information Matrix F is equal to negative expected Hessian of the log likelihood function and can be represented as 
The curvature pairs are computed every L steps and stored in $(S,Y)$ buffer only if they are sufficiently large. Further, adaQN performs a control condition by comparing the error at current and previous aggregated weights on a monitoring dataset. If the current error is larger than the previous error by a factor $\gamma$, the aFIM and curvature pair buffers are cleared and the weights are reverted to the previous aggregated weights. This heuristic though adds to the cost, avoids further deterioration of the performance due to noisy or stale curvatures.

\vspace{-3mm}
\section{Proposed Algorithm}
%\subsection{Nesterov's Accelerated Quasi-Newton (NAQ)}
\vspace{-1mm}

The Nesterov's Accelerated Quasi-Newton (NAQ) ~\cite{ninomiya2017novel} method achieves faster convergence compared to the standard quasi-Newton methods by quadratic approximation of the objective function at ${\bf w}_k+\mu {\bf v}_k$ and by incorporating the Nesterov's accelerated gradient $\nabla E({\bf w}_k+\mu {\bf v}_k)$ in its Hessian update. The update vector of NAQ can be written as
\vspace{-3mm}
\begin{equation}
{\bf v}_{k+1} = \mu_k {\bf v}_k + \alpha_k {\bf {g}}_k,
\vspace{-3mm}
\end{equation}
\vspace{-2mm} 
\begin{equation}
\vspace{-2mm}
{\bf {g}}_k=-{\bf {H}}_k \nabla E({\bf w}_k+\mu_k {\bf v}_k).
\end{equation}
\vspace{-2mm}
 The Hessian matrix ${\bf H}_{k+1}$ is updated using (5)
where
%\vspace{-1mm}
 \begin{equation}\label{eq:13}
{\bf s}_k = {\bf w}_{k+1} - ({\bf w}_k+ \mu{\bf v}_k), 
\end{equation}
\vspace{-5mm}
\begin{equation}
{\bf y}_k = \nabla E ( {\bf w}_{k+1} ) - \nabla E ({\bf w}_k+ \mu {\bf v}_k).
\end{equation} 
From (13) it can be noted that NAQ involves twice gradient calculation per iteration. 
In the limited memory form, LNAQ~\cite{LNAQ_shah} uses the last ${m_L}$ curvature pairs for the Hessian calculation. The curvature pairs that are used incorporate the momemtum and Nesterov's accelerated gradient term, thus accelerating LBFGS. Both NAQ and LNAQ are based on full batch and hence not suitable for solving large scale stochastic optimization problems.

\begin{algorithm}[tb]
\caption{ Direction Update}
\begin{algorithmic}[1]
\label{Algo:dirUp}
\Require current gradient ${\nabla E({\bf \theta}_k)}$, curvature pair (S,Y) buffer
\State $\tau = {\rm length}(S)$
\State ${\bf {\eta}}_k=\nabla E({\bf \theta}_k)$
\For {$i={\tau},\ldots,2,1~~$}
\State${\sigma}_i=({{\bf s}_i^{\rm T} {\bf {\eta}}_k})/({{\bf y}_i^{\rm T} {\bf s}_i})$
\State ${\bf {\eta}}_k={\bf {\eta}}_k-{ \sigma}_i {\bf y}_{i}$
\EndFor

\State ${\bf {\eta}}_k= {{\bf H}^{(0)}_k}{\bf {\eta}}_k$
%\State ${\bf {\eta}}_k={\bf {\eta}}_k({{\bf s}_k^{\rm T} {\bf y}_k}/{{\bf y}_k^{\rm T} {\bf y}_k})$

\For {$i:=1, 2, ..., {\tau~~}$}
\State${\beta}=({{\bf y}_i^{\rm T} {\bf {\eta}}_k})/({{\bf y}_i^{\rm T} {\bf s}_i})$
\State${\bf {\eta}}_k={\bf {\eta}}_k-({\sigma}_i-{\beta}){\bf s}_i$
\EndFor
\State ${\bf g}_k = -{\bf {\eta}}_k$
\State{\bf return}\;${\bf {g}}_k$
\end{algorithmic}
\end{algorithm}
 
\subsection{adaptive Stochastic NAQ  (aSNAQ)}
In this paper we propose a stochastic QN method by combining (L)NAQ and adaQN. The proposed method - adaptive Stochastic Nesterov Accelerated Quasi-Newton (aSNAQ) incorporates Nesterov's accelerated gradient term and a simple adaptively tuned momentum term. The algorithm is shown in Algorithm 2. 
aSNAQ also initializes ${\bf H}_k^{(0)}$ based on accumulated gradient information and uses aFIM for computing ${\bf y}$ for the Hessian computation as shown in (8) and (9) respectively. In aSNAQ, the gradient at ${\bf w}_k+\mu {\bf v}_k$ is used in the computation of the search direction while the gradient at ${\bf w}_{k+1}$ is used in the aFIM for Hessian approximation. Thus, aSNAQ also involves two gradient computations per iteration just like in NAQ. 
The curvature pairs are computed every L steps and stored in $(S,Y)$ only if sufficiently large. The momentum term $\mu$ is tuned by a momentum update factor $\phi$ as shown in step 22 of Algorithm 2.  aSNAQ also performs a error control check as shown in step 14-18. In addition to reseting the aFIM and curvature pair buffers and restoring old parameters, the momentum is also scaled down (step17). Thus there is adaptive tuning of the momentum parameter $\mu$ in the range $(\mu_{min},\mu_{max})$. Unlike adaQN the error control check is carried out on the same mini-batch. Further, direction normalization \cite{li2018implementation} is introduced in step 4 to improve stability and to solve the exploding gradient issue. 

\iffalse
\begin{equation}
[H_k^{(0)}]_{ii} = \frac{1}{\sqrt{ {\sum_{j=0}^{k} \nabla E({\bf w}_j+\mu{\bf v}_j )_i^2 } + \epsilon}} 
\end{equation} 
\fi

%In this paper we focus on proposing a stochastic QN method by combining (L)NAQ and adaQN. The proposed method - adaptive stochastic Nesterov Accelerated Quasi-Newton (aSNAQ) incorporates Nesterov's accelerated gradient term and a simple adaptively tuned momentum term. Like the adaQN method, aSNAQ also uses the accumulated Fisher Information matrix for computation of the curvature information and Hessian updates. In addition to that, we introduce direction normalization \cite{li2018implementation} to improve stability and to solve the exploding gradient issue. Thus the proposes method maintains a low-per iteration cost while accelerating the training process. The algorithm of the proposed aSNAQ method is shown in Algorithm 2.

%https://wiseodd.github.io/techblog/2018/03/11/fisher-information/

\begin{algorithm}[H]
    \centering
\caption{Proposed method - aSNAQ }
\begin{algorithmic}[1]
\label{Algo:adaNAQ}
\Require minibatch ${X_k}$, $\mu_{min}, \mu_{max}$, ${k_{max}}$, aFIM buffer {\it F} of size ${\it m_F}$ and curvature pair buffer $(S,Y)$ of size ${\it m_L}$ , momentum update factor $\phi$
\Ensure ${\bf w}_o$=${\bf w}_k\in \mathbb{R}^d$, $\mu$=$\mu_{min}$, ${\bf v}_k $,  ${\bf v}_o $ , ${\bf w}_s $, ${\bf v}_s$, ${k}$ and ${t=0}$
%\State $k \leftarrow 0$
%\State $t \leftarrow 0$
\While {$k < k_{max}$}
\State  {Calculate $\nabla E({\bf w}_k+\mu {\bf v}_k)$} 
\State Determine ${\bf g}_k$ using Algorithm 1
\State ${\bf g}_k = {\bf g}_k / ||{\bf g}_k||_2$  \algorithmiccomment{Direction normalization}

\State ${\bf v}_{k+1}\leftarrow \mu {\bf v}_k +\alpha_k {\bf {g}}_k$
\State ${\bf w}_{k+1}\leftarrow{\bf w}_k +{\bf v}_{k+1}$
\State  {Calculate $\nabla E({\bf w}_{k+1})$}  and store in $F$

\State {${\bf w}_s = {\bf w}_s + {\bf w}_{k}$}
\State {${\bf v}_s = {\bf v}_s + {\bf v}_{k}$}

\If {mod(k , L) = 0}
\State Compute average {${\bf w}_n = {\bf w}_s / L$} and {${\bf v}_n = {\bf v}_s / L$}
%\State {${\bf v}_n = {\bf v}_s / L$}
\State ${\bf w}_s = 0$ and ${\bf v}_s = 0$
\If {${t>0}$}
\If{$ E({\bf w}_n) > \gamma E({\bf w}_o)$}
\State {Clear $(S,Y)$ and $F$ buffers}
%\State {Clear aFIM buffer $F$}
\State Reset ${\bf w}_k = {\bf w}_o$ and ${\bf v}_k = {\bf v}_o$
\State Update $\mu ={\rm max}(\mu / \phi, \mu_{min})$ 
%\State ${\bf w}_k = {\bf w}_o$
\State {\bf continue}
\EndIf
\State {$ {\bf s } = {\bf w}_n - {\bf w}_o$}
\State {${\bf y} = \frac{1}{|{\it F}|}(\sum\limits_{i=1}^{|{\it F}|}{\it F_i} \cdot {\bf s})$}
\State Update $\mu ={\rm min}(\mu \cdot \phi, \mu_{max})$ 
\If {${\bf s}^T{\bf y} > \epsilon \cdot {\bf y}^T{\bf y}$}
\State Store curvature pairs ({\bf s},{\bf y}) in $(S,Y)$
%\State ${\bf w}_o={\bf w}_n$
%\State ${\bf v}_o={\bf v}_n$
\EndIf
%\Else
\EndIf

\State Update ${\bf w}_o={\bf w}_n$ and ${\bf v}_o={\bf v}_n$
%\State ${\bf v}_o={\bf v}_n$
\State $t \leftarrow t+1$
\EndIf
\State $k \leftarrow k+1$
\EndWhile
\end{algorithmic}
\end{algorithm}

\vspace{-7mm}
\section{Simulation Results}
\vspace{-1mm}
We study the performance of the proposed method on a toy example problem of sequence counting followed by MNIST classification problem. The simulations are performed using Tensorflow on  a simple one layer RNN network. Cross entropy loss function and  tanh activation is used.  We choose the aFIM buffer F size as $m_F=100$ and the limited memory size for the curvature pairs as $m_L=10$. The update frequency is chosen to be $L=5$, learning rate $\alpha = 0.01 $ and $\gamma=1.01$. The momentum update factor $\phi$ is set to 1.1. All weights are initialized with random normal distribution with zero mean and 0.01 standard deviation.% $\mathbb{N}(0,0.01)$. %with random normal distribution with zero mean and 0.01 standard deviation. 
%For all the problems in this paper, we study the performance on a simple one layer RNN network with tanh activation.% with 100 hidden neurons with a minibatch size $b=128$. 
%, $\mu_{min}$ is 0.1 and $\mu_{max}$ are 0.99 and 0.95 for regression and classification problems respectively. The momentum update factor is chosen to be $\phi=1.1$. Cross entropy loss function and tanh activation function is used.

%\subsection {Toy Example Problems}
\vspace{-1mm}
\subsection {Sequence Counting Problem}
\begin{figure}[htb]
\vspace{-0.4cm}
\begin{center}
\includegraphics[width=7cm]{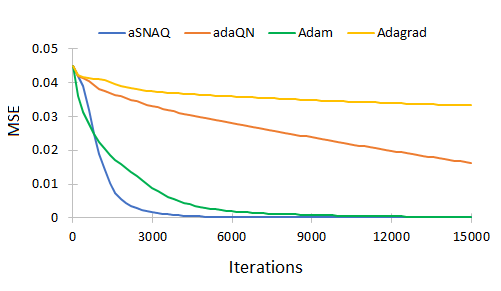}
\end{center}\vspace{-0.95cm}
\caption{MSE for sequence counting problem. }
\label{fig:count}
\vspace{-0.5cm}
\end{figure}

We evaluate the performance of the proposed algorithm on the sequence counting problem. Given a binary string (a string with just 0s and 1s) of length T, the task is to determine the count of 1s in the binary string. The number of hidden neurons is chosen to be 24, batch size ${b=50}$, ${T=20}$, $\mu_{min}=0.1$ and $\mu_{max}=0.99$. Fig \ref{fig:count} shows the mean squared error over 75 epochs. It can be observed that the proposed method clearly outperforms adaQN and Adagrad. On comparison with Adam, aSNAQ is faster in the initial iterations and becomes gradually close to Adam. 

\iffalse
\subsubsection {Adding Problem}
We evaluate the performance on the adding problem which is a sequence regression problem, designed to examine the power of recurrent models in learning long term dependencies \cite{le2015simple}.  The task is to obtain the sum of two numbers selected in a sequence of random signals with a uniform distribution in the range [0,1]. The input at each time step is a random signal and a mask signal indicating the two numbers to be added. Batch size $b=100$,  $\mu_{min}=0.1$ and $\mu_{max}=0.99$ and 100 hidden neurons are used.
Fig. \ref{fig:add} shows the mean squared error over 75 epochs 

\begin{figure}[htb]
\vspace{-0.3cm}
\begin{center}
\includegraphics[width=7cm]{D:/Spring19/RNN/count.png}
\end{center}\vspace{-0.7cm}
\caption{MSE for adding problem. }
\label{fig:add}
\vspace{-0.3cm}
\end{figure}
\fi
\subsection {Image Classification}
RNNs can be used to classify images by breaking the images into a sequence of pixel values. This can be done in two ways, namely row-by-row sequence and pixel-by-pixel sequence. In row-by-row sequencing at each timestep one row is fed as input while in pixel-by-pixel sequencing, at each timestep one pixel value is fed as input to the RNN in scanline order.

\iffalse
\subsubsection{Results on 8x8 MNIST Pixel by Pixel Sequence}
We first study the performance of the proposed method on a smaller reduced version of the MNIST dataset in which each sample is an 8x8 image representing a handwritten digit \cite{alpaydin1998optical}. Fig. \ref{fig:8x8} shows the training loss and accuracy of aSNAQ in comparison with adaQN and Adam.
\fi

\begin{figure}[htb]
\vspace{-3mm}
\begin{center}
\includegraphics[width=7cm]{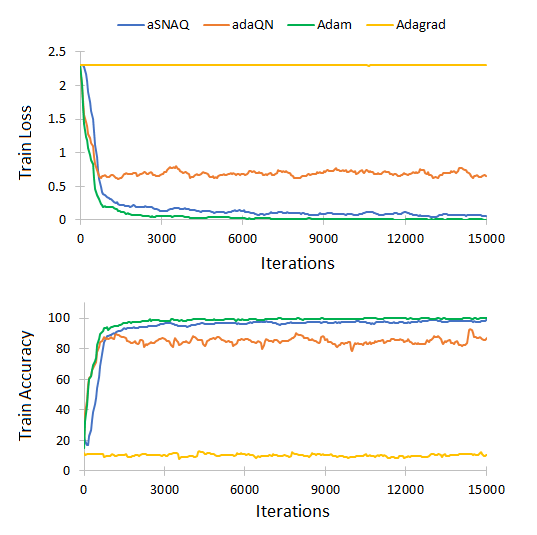}
\end{center}\vspace{-9mm}
\caption{Train loss and accuracy for 28x28 MNIST row by row sequence. }
\label{fig:scanner}
\vspace{-3mm}
\label{fig:rbr}
\end{figure}

\vspace{-0.3cm}
\subsubsection{Results on 28x28 MNIST Row by Row Sequence}
%\vspace{-0.2cm}
We study the performance of the proposed algorithm on the standard image classification problem MNIST. The input to the RNN is 28 pixels fed row-wise at each time step, with a total of 28 time steps. We choose batch size $b=128$,  $\mu_{min}=0.1$, $\mu_{max}=0.99$ and 100 hidden neurons. %The number hidden neurons used is 100.
 Fig. \ref{fig:rbr} shows the training loss and accuracy over 35 epochs. As seen from the results, we observe that Adagrad performs poorly while aSNAQ  performs better than Adagrad and adaQN and is almost on par with Adam.  

\begin{table}[htb]
\vspace{-3mm}
\begin{center}
\caption{Summary of Computational Cost. }\label{tab1}
\begin{tabular}{c c c}
\hline
\\[-3.2mm]
{Algorithm} &  {Computational Cost}\\ %& { Storage}\\
\hline
\\[-2mm]
%&&&&\multirow{4}{*}{\rotatebox{90}{full batch}} & 
{BFGS} &  ${nd + d^2 + \zeta nd}$\\% & ${d^2}$\\
{NAQ} &  ${2nd + d^2 + \zeta nd}$ \\%& ${d^2}$\\
\\[-3mm]
%{ LBFGS} &  ${nd + 4md + 2d + \zeta nd}$\\% & ${2md}$\\
%{LNAQ} &  ${2nd + 4md + 2d + \zeta nd}$\\% & ${2md}$\\
%\\[-3mm]
%{oBFGS} & ${2bd + d^2}$ \\
%{oNAQ} &  ${2bd + d^2}$\\
%\\[-3mm]
%{oLBFGS} &  ${2bd + 6md}$ \\
%{oLNAQ} &  ${2bd + 6md}$ \\
%\\[-3mm]
{ adaQN} &  ${bd + (4m_L+ m_F + 2)d + (b+4)d/L}$ \\%& ${(2m_L+m_F)d}$\\
{aSNAQ} &  ${2bd + (4m_L + m_F + 3)d + (b+4)d/L}$ \\%& ${(2m_L+m_F)d}$\\
\\[-3mm]
\hline
\end{tabular}
\end{center}
\vspace{-2mm}
\end{table}

\begin{figure}[]
\vspace{-0.2cm}
\begin{center}
\includegraphics[width=7cm]{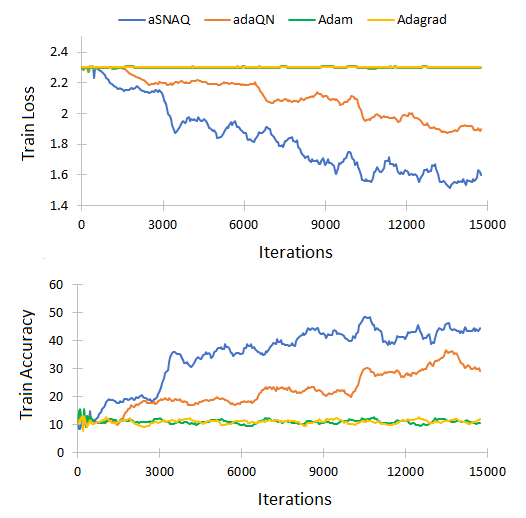}
\end{center}
\vspace {-8mm}
\caption{The average train loss and accuracy for 28x28 MNIST pixel by pixel sequence. }
\vspace {-3mm}
\label{fig:pbp}
\end{figure}
\vspace{-0.3cm}
\subsubsection{Results on 28x28 MNIST Pixel by Pixel Sequence}
\vspace{-0.1cm}
We further extend to study the performance of the proposed algorithm using pixel-by-pixel sequence. The pixel-by-pixel sequence based classification is a hard task since the model has to keep a very long-term memory. It involves 784 time steps and is a much harder problem compared to the regular classification methods. Fig.\ref{fig:pbp} shows the training loss and accuracy over 35 epochs. In pixel by pixel sequence, both Adam and Adagrad methods perform poorly. Though the overall training accuracies are low with this simple one-layer RNN,  aSNAQ show significant improvement in training compared to adaQN, Adam and Adagrad.

%\subsection{Computational Cost}
% The computation cost is given in Table 1. The typical second order methods such as BFGS method incur a cost of $nd + d^2 + \zeta nd$ in gradient, Hessian and linesearch compuation respectively. In case of NAQ, an additional $nd$ cost is incured due to twice gradient compuation.  adaQN and the proposed aSNAQ being stochastic methods, the computation cost in gradient calculation is $bd$ where b is the minibatch size. The Hessian approximation is carried out using the aFIM and two-loop recursion, thus reducing the computation cost to $(4m_L+m_F+2)d$. Further the error control check adds to an additional $(b+4)d/L$. aSNAQ has an additional cost $bd$ and $d$ due to twice gradient computation and direction normalization.  The storage cost of BFGS and NAQ is $d^2$ while adaQN and aSNAQ is ${(2m_L+m_F)d}$.

\vspace{-0.1cm}
\subsection{Conclusion}
\vspace{-0.05cm}
In this paper we have proposed an adaptive stochastic Nesterov's accelerated quasi-Newton method. The computation cost is given in Table 1. The typical second order methods such as the BFGS method incurs a cost of $nd + d^2 + \zeta nd$ in gradient, Hessian and linesearch compuation respectively, where $n$ = $|T_r|$. In case of NAQ, an additional $nd$ cost is incured due to twice gradient compuation.  The adaQN and proposed aSNAQ methods being stochastic methods, the computation cost in gradient calculation is $bd$ where b is the minibatch size. The Hessian approximation is carried out using the aFIM and two-loop recursion, thus reducing the computation cost to $(4m_L+m_F+2)d$. Further the error control check adds to an additional $(b+4)d/L$ computations. aSNAQ has an additional cost of $bd$ and $d$ due to twice gradient computation and direction normalization.  The storage cost of BFGS and NAQ is $d^2$ while adaQN and aSNAQ is ${(2m_L+m_F)d}$. The proposed method attempts to incorporate second order curvature information while  maintaining a low per iteration cost. Incorporating the Nesterov's accelerated gradient term has shown to improve the performance in  the training of RNNs compared to adaQN and other first order methods.  Further analysis of the proposed algorithm on other RNN structures such as LSTMs and GRUs with bigger sequence modeling problems along with convergence property analysis are left for future work.
\vspace{-0.2cm}
\bibliography{references}{}
\bibliographystyle{splncs}

\iffalse

\fi

\end{document}